\documentclass{article}
\usepackage{ijcai21}

\usepackage{times}
\usepackage{soul}
\usepackage{url}
\usepackage[hidelinks]{hyperref}
\usepackage[utf8]{inputenc}
\usepackage[small]{caption}
\usepackage{graphicx}
\usepackage{amsmath}
\usepackage{amsthm}
\usepackage{booktabs}
\usepackage{algorithm}
\usepackage{algorithmic}
\usepackage{comment}
\usepackage{setspace}
\usepackage{hhline}
\usepackage{float}
\usepackage{subfig}
\usepackage{amssymb}
\usepackage{graphicx}
\usepackage{amsmath}
\usepackage{natbib}
\usepackage[title]{appendix}

\title{Relevance-Guided Modeling of Object Dynamics for Reinforcement Learning}

\author{
William Agnew\footnote{Contact Author}\and
Pedro Domingos\
\affiliations
University of Washington School of Computer Science and Engineering\\
\emails
\{wagnew3, pedrod\}@cs.washington.edu
}

\begin{document}

\maketitle

\begin{abstract}
  Current deep reinforcement learning (RL) approaches incorporate minimal prior knowledge about the environment, limiting computational and sample efficiency. \textit{Objects} provide a succinct and causal description of the world, and many recent works have proposed unsupervised object representation learning using priors and losses over static object properties like visual consistency. However, object dynamics and interactions are also critical cues for objectness. In this paper we propose a framework for reasoning about object dynamics and behavior to rapidly determine minimal and task-specific object representations. To demonstrate the need to reason over object behavior and dynamics, we introduce a suite of RGBD MuJoCo object collection and avoidance tasks that, while intuitive and visually simple, confound state-of-the-art unsupervised object representation learning algorithms. We also highlight the potential of this framework on several Atari games, using our object representation and standard RL and planning algorithms to learn dramatically faster than existing deep RL algorithms.
\end{abstract}

\section{Introduction}

Deep RL has achieved impressive performance in many environments (\cite{mnih2015human, silver2017mastering}). 
However, the sample inefficiency of deep RL algorithms limits their applicability, especially to domains where experiences are expensive. In addition, learned policies lack robustness to changes in visual inputs.
In many environments, humans can achieve good performance after far less experience than machines, pointing to human-inspired priors as a means of improving sample efficiency. Perceiving the world in terms of objects, or groups of percepts that maintain properties across time, is a fundamental aspect of human perception and cognition (\cite{spelke1990principles}). Past work in RL has considered perceiving the world in terms of objects as humans do (\cite{diuk2008object, scholz2014physics}). However, obtaining object representations without supervision remains a challenging problem. \cite{naddaf2010game}, \cite{liang}, and \cite{machado2017revisiting} begin to tackle the problem of recognizing objects without extensive supervision or hand crafted, object specific recognition algorithms by introducing Blob-PROST, BASS, and DISCO, large object feature sets. Many recent works have made progress on this front by using information bottlenecks, such as encoder-decoder networks, to reconstruct visual observations (\cite{goel2018unsupervised, zhu2018object, anand2019unsupervised, kulkarni2019unsupervised, greff2019multi, veerapaneni2019entity, lin2020space, du2020unsupervised}). Since objects have consistent visual properties, these frameworks naturally learn disentangled object representations as an information-efficient encoded representation. 

Objects have not only consistent appearance but also consistent dynamics and reward behavior. Using dynamics to determine objectness allows the agent to ignore background visual distractors and successfully associate object parts with different appearances. Using reward and value to determine objectness allows for learning \textit{task-relevant} object representations by allowing for ignoring objects with no impact on value and reward. Taken together, adding dynamics and reward cues enables the learning of more succinct object representations and provides additional signals to facilitate learning of object representations in fewer samples. Recently, \cite{du2020unsupervised} have added a single-step, contact-free dynamics model to visual-loss-based object discovery networks, and \cite{creswell2020alignnet} learn a dynamics model to track a fixed, known set of objects, making progress towards incorporating dynamics-based notions of objectness.

In this paper we present an unsupervised framework for rapidly learning object representations using visual, dynamics, and reward cues. Our framework extends existing work both by incorporating reward information when determining objectness and by permitting the use of more powerful dynamics model architectures, allowing the modelling of objects over many timesteps and under contacts and agent actions, two vital dynamics cues for determining objectness. Our key insight is that objects have similar properties across time; therefore, the observed dynamics and rewards of a  correctly inferred set of objects will exhibit low distribution shift and be easier to model accurately. Inversely, if a proposed mapping of observations to objects is inaccurate, then it will not have consistent properties across time and the observed dynamics and reward will exhibit distribution shift and be harder to model accurately. We use this idea to create an \textit{object objective} that evaluates a proposed object representation based on the accuracy of learned dynamics, reward, and value object models. By optimizing this objective we show we can learn object representations that are both more compact and learnable after hundreds, as opposed to tens or hundreds of thousands, of observations. Finally, we develop an explicit object representation in terms of object positions, velocities, accelerations, and contacts and use this representation for our RL.

\begin{figure}
\centering
\includegraphics[width=0.48\textwidth]{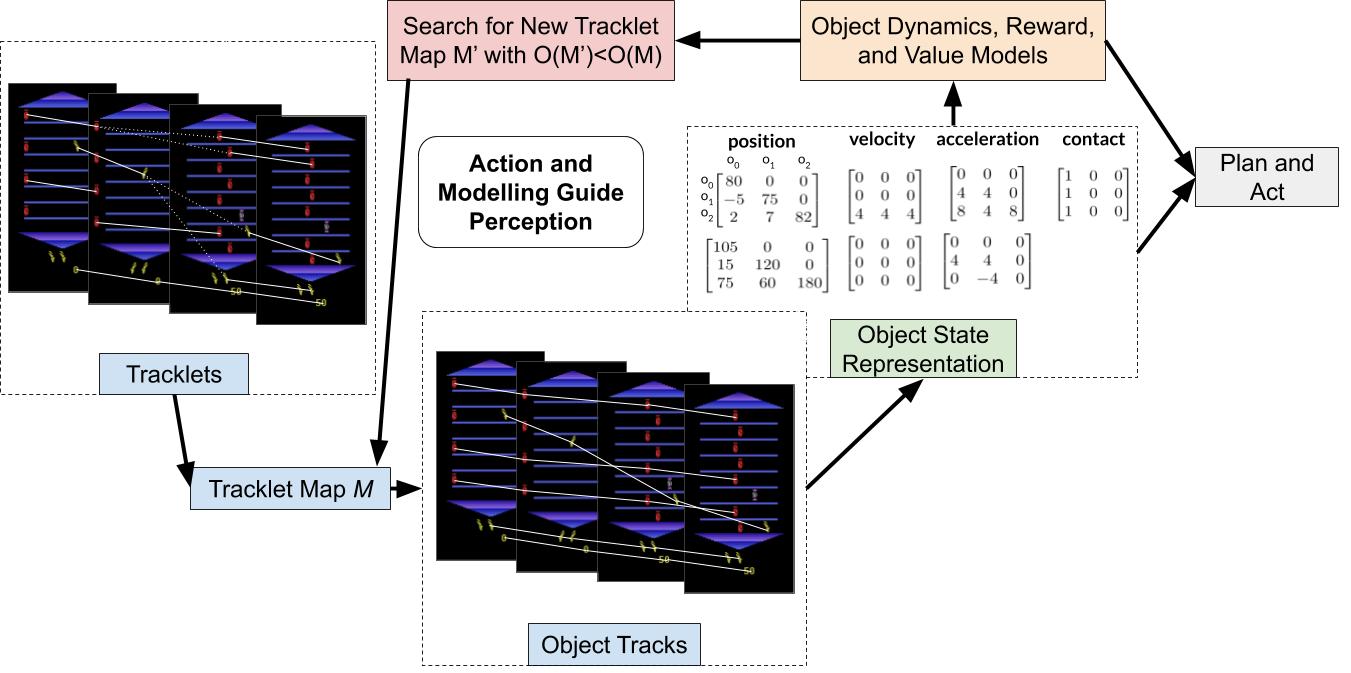}
\caption{Diagram of OLRL framework. a) tracklets and mapping: blue. b) object state representation: green. c) object modelling: orange. d) tracklet map inference: red. e) object-centric learning and planning: grey.}
\label{fig:OLRL_overview}
\end{figure}

Learning causality in sparse reward settings is a fundamental challenge in reinforcement learning. We use another human object prior to address this problem: objects interact only when in contact (\cite{spelke1990principles}). We implement this prior by defining an object-contact state representation and reward-and-value approximation architecture. We represent the relationship between two objects by their relative position, velocity, acceleration, and whether they are in contact. This representation is both compact and sparse. We model reward and estimated value as a function of the pairwise features between contacting objects, enforcing the notion that objects interact only upon contact.

We combine our object inference, object state representation, and object dynamics, value, and reward models to create an Object-Level Reinforcement Learner (OLRL) framework (Figure \ref{fig:OLRL_overview}). We compare OLRL with deep RL algorithms on MuJoCo (\cite{todorov2012mujoco}) and Atari (\cite{machado2017revisiting}) environments, and we show that it significantly improves sample and computational efficiency. Furthermore, we compare OLRL to humans learning to play Atari games and find OLRL achieves slightly better sample efficiency. Currently deep RL research can be resource and time intensive due to the vast number of experiences needed for training. Our object detection and tracking model can serve as a general preprocessor for RL, greatly increasing the pace and sample efficiency of experiments.

In summary, contributions include an OLRL framework that encompasses:
\begin{itemize}
  \item An unsupervised object discovery objective \textit{objective function} which encourages inference of objects with consistent dynamics, reward, and value behavior (Figure 1.a-c)
  \item A flexible \textit{reward and value estimation architecture} which encodes the causality of object contacts (Figure 1.d)
  \item An \textit{object-centric RL agent} which is significantly more sample efficient than standard deep RL algorithms (Figure 1.e)
\end{itemize}

\section{Background}
In this paper we consider solving MDPs with state represented by pixels. Formally, at each time $t \in [0,T]$, our agent observes an RGBD image, $s \in \mathbb{R}^{H \times W \times 4}$, takes an action $a^t \in A$, and receives stochastic reward $R(s^t, a^t) \in \mathbb{R}$. The MDP state then transitions as given by $G(s^{t+1}, s^t, a^t)=P(s^{t+1}|s^t, a^t)$, a stochastic transistion function. Given a discount factor $\gamma \in [0, 1]$, the agent's goal at each timestep is to take the action that maximizes the discounted reward, or value: $\sum_{t=0}^T R(s^{t}, a^{t}) \gamma^{t}$. Reinforcement learning is one popular family of solutions to this problem, solving the MDP by learning an optimal state-action value function $Q_{*}(s, a)=\max_{\pi} Q_{\pi}(s, a)$ for all $s \in S$ and $a \in A$, and following an optimal policy $\pi$ that is greedy with respect to the optimal state-action value function. Model-based RL approaches learn models of the MDP transistion and reward functions to use either as a simulator for learning $Q^*$ or for planning actions.

Substantial work has addressed model-based deep RL and predicting future states or values with deep networks. \cite{oh2015action} and \cite{chiappa2017recurrent} use deep neural networks to accurately predict Atari frames hundreds of steps into the future. However, these works rely on hundreds of thousands of training frames and computationally intensive deep architectures, do not consider environment stochasticity, and predict at the pixel level rather than the more efficient object level. \cite{kaiser2019model} integrate similar pixel-level frame prediction for Atari into a deep RL agent. \cite{ebert2017self} and \cite{higuera2018synthesizing} learn to predict future state, but they still predict either very low-level features that are difficult to learn on or high-level features that require many samples to model effectively. 

A fundamental problem of RL, especially in the visual domain, is the size of the state space, which can greatly complicate learning and generalization and increase the number of experiences needed to learn. One classic method of addressing this problem is using objects, or groups of percepts with consistent properties across time. By representing the state in terms of objects, the state space dimensionality can be greatly reduced, allowing more sample efficient and robust learning. Object state representations for RL were proposed by \cite{diuk2008object} with a formal framework for describing and reasoning about object interactions. Unlike our object recognition algorithm, these techniques require environment-specific object labels. In the pixel-based MDPs studied in this paper, objects are present as clusterings of pixels across timesteps. Assume the agent observes $T$ sequential RGBD observations of a $d$-dimensional environment as four-channel images $I \in \mathbb{R}^{T \times H \times W \times 4}$. These observations may be divided into a set of \textit{tracklets} $L$, where each tracklet $l \in \left\{0,1\right\}^{T \times H \times W}$ is a set of instance masks (1 denoting part of the instance, 0 not). Two tracklets can be combined by taking the union of their instance masks at each timestep: $l \cup l'=u$ where $l,l',u \in \left\{0,1\right\}^{T \times H \times W}$. In this way we may combine several tracklets, intuitively representing parts of a particular object, into a single track corresponding to the whole object.

\section{Model}
We define objects at a high level as entities that have consistent properties across time. We determine objectness by searching for groups of percepts that have not just consistent visual properties but also consistent behavior: dynamics, reward, and value. We now describe OLRL, a new model for learning object representations without supervision. Our model works by first learning predictive models of the dynamics, reward, and value of the tracklets in terms of their features (Figure 1.b-c). Next, it combines tracklets together by searching for tracklet combinations that reduce the overall error of the dynamics, reward, and value models (Figure 1.d). Finally, it uses the learned object dynamics, reward, and value models to plan optimal actions to take (Figure 1.e).

\subsection{Object Representation}
Given a set of tracklets $L$, we represent the state at time $t$, $o^t$, by a tuple $(p_a,p_r,v_a,v_r,a_a,a_r,c)$ where $p_a, v_a, a_a \in \mathbb{R}^{|L|}$ are the absolute positions, velocities, and accelerations of objects, respectively, and $p_r, v_r, a_r \in \mathbb{R}^{|L| \times (|L|-1)/2}$ are the relative positions, velocities, and accelerations between each pair of objects, respectively. Finally, $c \in \left\{0,1\right\}^{|L| \times (|L|-1)/2}$ encodes contacts between object pairs. We compute object position by taking the median mask pixel position, and object contact by searching for object masks with adjacent pixels. This object state representation succinctly represents the high-dimensional pixel observations as a set of intuitive and casual features, allowing for more efficient learning.

\subsection{Model Learning}
We learn predictive models of dynamics, reward, and value for each tracklet. We can train these models with direct supervision, allowing our agent to learn accurate estimators with few samples. In section 3.4 we use these models as a guide for determining which tracklets to combine and for planning which actions to take. Let $D: o^t, a^t \rightarrow [0,1]^{d \times |K| \times |V_d|}$ be a probabilistic object velocity model over a discrete set of velocities $V_d$. Let $R: o^t, a^t \rightarrow r^t$ be an agent reward model, and let $V: o^t, a^t \rightarrow v^t$ be an agent value model. Then let $E_D(o^{\hat{T}}, a^{\hat{T}})$, $E_R(o^{\hat{T}}, a^{\hat{T}})$, and $E_V(o^{\hat{T}}, a^{\hat{T}})$ be respective model errors on some held-out validation times $\hat{T}$:

\[E_D(o_M^{\hat{T}}, a^{\hat{T}})=\frac{1}{|\hat{T}||T|}\sum_{t \in \hat{T}} D(o_M^{t}, a^{\hat{T}})\cdot |o_M^{t+1}[v_a]-V_d|, \]

where \noindent $o_M^{t+1}[v_a]$ is a $d \times |T|$ matrix of absolute velocities in the $t+1$th observation,

\[E_R(o_M^{\hat{T}}, a^{\hat{T}})=\frac{1}{|\hat{T}|}\sum_{t \in \hat{T}}(R(o_M^{t}, a^{\hat{T}})-r^t)^2 \]
\[E_V(o_M^{\hat{T}}, a^{\hat{T}})=\frac{1}{|\hat{T}|}\sum_{t \in \hat{T}}(V(o_M^{t}, a^{\hat{T}})-v^t)^2. \]

\subsection{Object Inference}
To obtain a minimal object representation, we seek to infer a map from tracklets to tracks, $M: l \in L \rightarrow k \in K$, where $k=\bigcup_{l:M(l)=k} l$. Let $o_M$ be the object state under tracklet map $M$. Intuitively, because objects are things we consistent dynamics, reward, and value across time, a good object map will have low distribution shift and allow for learning of accurate dynamics, reward, and value models, and vice versa. We formalize this by defining an \textit{object objective}, which measures the quality of a particular object map:

\[O(M)=E_D(o_M^{\hat{T}}, a^{\hat{T}})+E_R(o_M^{\hat{T}}, a^{\hat{T}})+E_V(o_M^{\hat{T}}, a^{\hat{T}})\] 

By minimizing $O(M)$, we find the object representation that allows us to learn the most accurate object models. Solving this problem exactly is computationally expensive, as there are exponentially many ways to map a set of tracklets into tracks. In practice, we take a greedy approach: we randomly select two tracklets, merge them together for form map $M'$, and accept $M'$ if $O(M')<O(M)$, and find that this approximation is sufficient to produce succinct representations.

Our system uses off-the-shelf tracking and segmentation algorithms to preprocess $I$ into a set of initial tracklets $L$. The off-the-shelf tracking and segmentation algorithms are tuned to oversegment and undertrack, so each tracklet contains at most one object with high confidence, but objects may be split across multiple tracklets.

\begin{figure}
\centering
\includegraphics[width=0.5\textwidth]{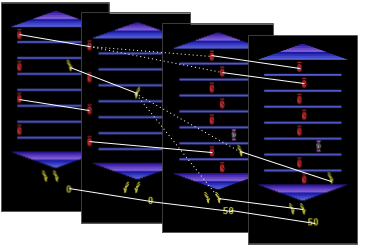}
\caption{Object tracklets (white lines) over percepts and proposed object tracklet merges (dotted white lines) over four successive environment observations. Our object representation objective scores proposed object tracklet merges, allowing us to accept the correct merges.}
\label{fig:asterix_tracklets}
\end{figure}

\subsection{Object-Centric Reinforcement Learning}
We propose a model-based RL architecture motivated by the observation that objects tend to act on each other only when in contact (\cite{spelke1990principles}). We model value and reward as functions of contacting objects and their relative features rather than learning to predict reward and value as a function of all object features, further reducing the dimensionality of our object representation. For computational efficiency, we consider only pairwise contacts. For an object pair $i, i'$, we define their pairwise features $o_{i,i'}$ as the relative position, velocity, and acceleration between $i$ and $i'$ and the absolute position, velocity, and acceleration of $i$ and $i'$. Let $c(i,i')=1$ if $i$ and $i'$ make contact in the current observation and $n(i,i')=\sum_{i,i' \in M} c(i,i')$. Then, we model reward and value as 
\[R(o,a)=\frac{1}{n(i,i')}\sum_{i,i' \in M, c(i,i')=1} R_{i,i'}(o_{i,i'},a)\] 
\[V(o,a)=\frac{1}{n(i,i')}\sum_{i,i' \in M, c(i,i')=1} V_{i,i'}(o_{i,i'},a)\] 
Since we take the mean of the pairwise models, each can be independently trained to predict the observed rewards and estimated state-action values, enhancing flexibility when choosing specific learning algorithms for each $R_{i,i'}$ and $V_{i,i'}$.

Our agent can then use the learned dynamics, reward, and value models to plan into the future and choose the action sequence with the highest predicted value. For a plan $p=\left\{ o^0, o^1, ... \right\}$ consisting of a set of predicted object states and corresponding action sequence $s_a=\left\{a_0, a_1, ...\right\}$, the value is $Q(p,s_a)=\gamma^{|p|} V(o^{|p|}, a^{|p|})+\sum_{t \in |p|} \gamma^t R(o^t, a^t)$.

\subsection{Implementation}
\textit{Segmentation and Tracking}. We use UIOS (\cite{xie2019best}), a pretrained RGDB instance segmentation network, to produce initial segmentations. To form initial tracklets, we match segments in adjacent observations by assigning each segment pair a score based on visual similarity and physical proximity (see the Appendix for more details). We then greedily assign pairs with the lowest score below a certain threshold. Any unmatched segmentation is added as a new tracklet.

\textit{Object Objective Function}. We model object dynamics by learning a probabilistic model of velocity. We first discretize velocity into $V_d=\left\{-30, ... 30\right\}$. We then use an XGBoost tree (\cite{chen2016xgboost}) for each object in each dimension to predict a probability vector over possible velocities for the object in the next time step. We choose a discrete model architecture over a continuous one to best model the often discontinuous outcomes of collisions, for example, elastic vs. perfectly inelastic behavior. We select a random 20\% of observations to hold out for error computation as $\hat{T}$.

\textit{Model-Based RL}. We use on-policy Monte Carlo RL (\cite{sutton2018reinforcement}) to compute $Q(s, a)$ targets with $\gamma=0.95$. Each time a new tracklet map is accepted, the agent retrains the dynamics, reward, and value models on the new object representation over past experiences. Actions are selected by planning two actions into the future and selecting the action that maximizes predicted expected value. Since our dynamics model outputs a probability over future states, we sample ten possible future states for each action path and take their mean value.

\section{Experiments}
We first evaluated our approach on a set of 3D environments with challenging confounding object behavior. We implemented these environments in MuJoCo (\cite{todorov2012mujoco}), a fast and realistic physics engine. Next we ran extensive ablation studies on the object inference, pairwise reward and value modelling, and learning and planning components of OLRL. Then, we applied our object-centric agent to several Atari environments and showed that it not only learned significantly faster than state-of-the-art sample efficient deep RL algorithms, but that it also learned slightly faster than humans.

\begin{figure}
\centering
\includegraphics[width=0.45\textwidth]{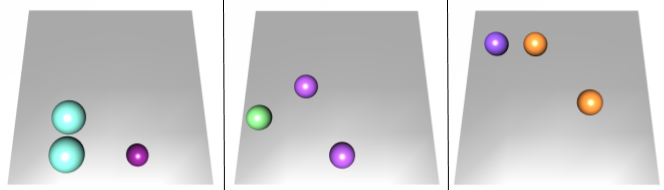}
\caption{Starting configurations of three successive episodes of the tabletop MuJoCo tasks. The agent must control one ball to either gather or avoid the other balls.}
\label{fig:mj_random}
\end{figure}

\textbf{MuJoCo Environments}. We proposed a suite of custom 3D MuJoCo environments to highlight current challenges with learning object representations. The base tasks were controlling a tabletop agent to either gather or avoid moving objects, implemented by giving the agent a reward of +1 and -1, respectively, for contacting these objects. These tasks were deliberately quite simple, and easily solved by modern deep RL algorithms or segmented by current unsupervised segmentation algorithms. We introduce two \textit{object randomization} tasks, gather+r and avoid+r (Figure~\ref{fig:mj_random}), where we randomize the color, size, and starting positions of the both the agent and reward objects after each episode, simulating changes in visual conditions or the objects themselves. This object appearance randomization requires using features beyond just visual appearance to successfully segment and track objects, challenging existing models.

\begin{figure}
\begin{center}
\scriptsize
\begin{tabular}{| c |  c | c | c | c |}
\hline
 Algorithm & gather & avoid & gather+r & avoid+r \\
\hhline{|=|=|=|=|=|}
 A2C & 35.3$\pm$ 4.2 & \textbf{-12.0$\pm$ 3.7} & 38.0$\pm$ 6.2 & -15.7$\pm$ 9.3\\ 
 \hline
 OLRL & \textbf{103.3$\pm$ 41.8} & -17.7$\pm$ 13.7 & \textbf{62.0$\pm$ 3.7} & \textbf{-8.5$\pm$ 7.3}\\
 \hline
 OLRL-m  & 76.0$\pm$ 55.0 & -13.0$\pm$ 6.0 &  34.7$\pm$ 16.8 & -25.3$\pm$ 24.1 \\
 \hline
 Random & 29$\pm$ 3.7 & -20$\pm$ 7.3 & 29.3$\pm$ 7.6 & -32.3$\pm$ 15.0\\
 \hline
\end{tabular}
\end{center}
\caption{Mean final episode scores on MuJoCo environments, with standard deviations. A2C trained for 100,000 steps; OLRL trained for 2000.}
\label{fig:mj_results}
\normalsize
\end{figure}
\normalsize

\begin{figure}
\centering
\includegraphics[width=0.45\textwidth]{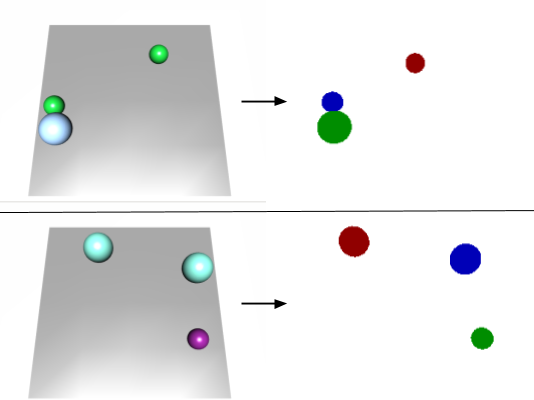}
\caption{Left: RGB observation, Right: object segmentation. The top row shows segmentation at time t=10, and the bottom at t=450. By considering object dynamics and behavior, OLRL can together map the agent-controlled objects, light blue on top and purple on bottom, enabling transfer of past experiences. }
\label{fig:mj_seg}
\end{figure}

Figure~\ref{fig:mj_results} compares the performance of our algorithm, OLRL, with A2C (\cite{wu2017scalable}) and an agent that takes random actions. In addition, we show OLRL without tracklet mapping, OLRL-m. OLRL and OLRL-m outperform A2C on the gather and avoid tasks with only 1/50th the experiences using our object representation and contact-centric reward and value estimation architecture. For the gather+r and avoid+r tasks, both A2C and OLRL-m performed little better than random due to object appearance being randomized each episode, which prevented visual-based transfer of learned object properties from one episode to the next, essentially forcing these algorithms to start learning from scratch each episode. However, OLRL, by using our object objective function to reason about the behavior of objects, quickly matched new objects with past objects and associated value, reward, and dynamics models, as shown in Figure~\ref{fig:mj_seg}. 

\textbf{Ablations}:
Figure \ref{fig:ablations} examines the impact of different components of OLRL on performance. We evaluated the quality of our object representation by replacing our object inference with SPACE (\cite{lin2020space}), a unsupervised object segmentation network, in \textit{SPACE+OCRL}. We determined the importance of using object features, rather than just object segmentations, by comparing to A2C trained using UIOS segmentations as input in \textit{UOIS+A2C}. We ablated our pairwise ensemble of reward and value estimators in \textit{Unary Approx.} where we used only one estimator to predict reward and value. Finally, in \textit{OLRL+A2C}, we input our object features to A2C to ablate the importance of our object-centric reinforcement learning. These ablations demonstrate the effectiveness of each component of OLRL. The poor performance of \textit{UOIS+A2C} and \textit{OLRL+A2C} show that object segmentations or representations alone are insufficient for fast learning in this challenging environment, as A2C is unable to learn much with them. While \textit{Unary Approx.} acheives the best performance of the ablations, it still performs worse than using object pairwise reward and value estimators as in OLRL. \textit{SPACE+OCRL} highlights the robustness of OLRL to both object appearance randomization, which occurs in the gather+r and avoid+r environments, and to objects vanishing or moving large distances, which occurs in the gather and avoid environments. By defining objects as having consistent motion, value, and reward in our object objective, OLRL is robust to large changes in appearance and motion, pointing to better generalization to novel objects and sim-to-real or other transfer.

\begin{figure}
\begin{center}
\scriptsize
\begin{tabular}{| c |  c | c | c | c |}
\hline
 algorithm & gather & avoid & gather+r & avoid+r \\
 \hhline{|=|=|=|=|=|}
 UOIS+A2C & 30.3$\pm$11.1 & -18.7$\pm$4.4 & 32.0$\pm$9.0 & -26.0$\pm$5.0 \\ 
 \hline
 Unary Approx. & 44.5$\pm$3.5 & \textbf{-15.0$\pm$7.0} & 38.5$\pm$2.5  & -14.5$\pm$10.5\\
 \hline
 SPACE+OCRL  & 32.0$\pm$4.0 & -20.7$\pm$10.2 &  26.3$\pm$8.4 & -12.7$\pm$2.4 \\
 \hline
 OLRL+A2C &  19.0$\pm$5.0 & -29.3$\pm$8.2 & 28.0$\pm$1.0 & -23.5$\pm$2.5\\
 \hline
 OLRL & \textbf{103.3$\pm$41.8} & -17.7$\pm$13.7 & \textbf{62.0$\pm$3.7} & \textbf{-8.5$\pm$7.3}\\
 \hline
\end{tabular}
\end{center}
\caption{Ablations of different parts of OLRL. A2C trained for 100,000 steps, OLRL trained for 2000.}
\label{fig:ablations}
\normalsize
\end{figure}

\textbf{Atari Experiments}: The Atari games we chose present many challenging and general object inference problems, including distractor objects, large changes in orientation or observed size due to occlusion, visually identical but behaviorally different objects, and discontinuities in observed dynamics. Successfully segmenting, tracking, and succinctly representing these objects requires difficult reasoning over dynamics, reward, and value, which humans can do with ease, highlighitng the importance of this reasoning its and generalizability beyond Atari. First, we examined the object representations OLRL learns for each Atari game and noted several challenges OLRL overcame by reasoning about dynamics and reward. Then, we compare the performance of OLRL to deep RL algorithms; and finally, we compare OLRL's learning speed to that of human players. We used Felzenszwalb segmentation (\cite{felzenszwalb2004efficient}) rather than UOIS and treated all points as being at depth zero for our Atari experiments.


\begin{figure}
\centering
\includegraphics[width=0.23\textwidth]{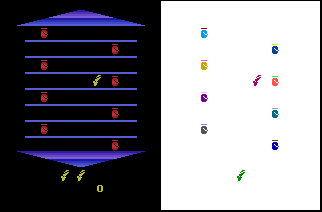}
\includegraphics[width=0.23\textwidth]{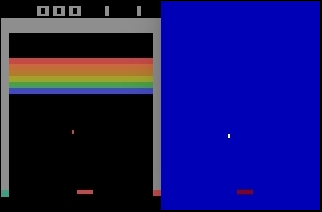}
\includegraphics[width=0.23\textwidth]{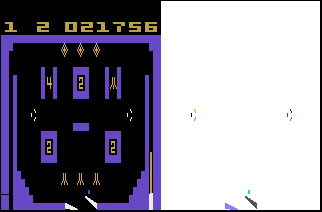}
\includegraphics[width=0.23\textwidth]{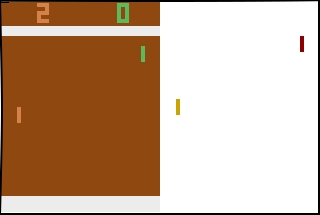}
\caption{Object-level representations of Atari games. Clockwise from top left: Asterix, Breakout, Pong, and Video Pinball. Left: observed Atari game frames. Right: Inferred OLRL object segmentations.}
\label{fig:objectlevelrep}
\end{figure}

\textbf{Object Inference Challenges}. Using our object objective function to combine map tracklets together, we rapidly inferred succinct object representations for each game, as shown in Figure~\ref{fig:objectlevelrep}. Each game contains many \textit{distractor objects}, which are visually distinct and even move but are of little use for learning a policy. Specifically, each game has many colorful static background objects. More challengingly, each game has a numerical score, and each numeral frequently appears and disappears, exhibiting distinct dynamics from the background. In three of the games, Asterix, Breakout, and Pong, the agent-controlled object frequently undergoes \textit{changes in orientation or observed size} due to occlusion, rotating up to 180$^\circ$ or doubling or halving in size between successive observations, posing a challenge for any purely visual-based tracking algorithm.  Finally, two games, Asterix and Pong, have task-irrelevant \textit{objects that are visually identical} or near-identical to the agent-controlled object. This challenge is compounded by frequent \textit{discontinuities in observed dynamics} when objects teleport large distances to starting positions during game resets: physical proximity and visual similarity together are insufficient to track them. OLRL correctly segmented and tracked objects through these challenges by relying on behavior to associate tracklets, ignoring task-irrelevant objects or associating visually distinct but behaviorally similar percepts.

\begin{figure}
\begin{center}
\scriptsize
\begin{tabular}{| c |  c | c | c | c |}
\hline
 Algorithm & Asterix & Breakout & Pong & Video Pinball \\
 \hhline{|=|=|=|=|=|}
 OLRL & \hspace{-5pt} 4250$\pm$250 & \hspace{-5pt} \textbf{136.7$\pm$68.9} & \hspace{-5pt} -13.3$\pm$0.4 & \hspace{-5pt} \textbf{16795.7$\pm$557.6}\\ 
 \hline
 DQN & \hspace{-5pt} 2866.8$\pm$1354.6 & \hspace{-5pt} 35.1$\pm$22.6 & \hspace{-5pt} \textbf{15.1$\pm$1.0} & \hspace{-5pt} 15,398.5$\pm$2,126.1\\
 \hline
 Blob-PROST  & \hspace{-5pt} \textbf{4358.0$\pm$431.6} & \hspace{-5pt} 20.2$\pm$1.9 &  \hspace{-5pt} 14.5$\pm$2.0 & \hspace{-5pt} 13,398.0$\pm$3,643.7 \\
  \hline
 SimPLE  & 1128.3 & 16.4 &  12.8 &  -- \\
  \hline
 SPR  & 997.8 & 17.1 &  -5.9 &  -- \\
 \hline
\end{tabular}
\end{center}
\caption{Atari learning results.}
\label{fig:RLcomp}
\normalsize
\end{figure}

\textbf{Comparison to Deep RL}. Figure~\ref{fig:RLcomp} compares OLRL to DQN, Blob-PROST, and two recent sample-efficient deep RL agents for Atari, SimPLe (\cite{kaiser2019model}) and DER (\cite{schwarzerdata}). DQN and Blob-PROST were trained for 100 million steps, SimPLe and SPR for 100,000, and OLRL for 10,000 steps. Our object-level agent reached similar or greater performance than DQN and Blob-PROST with approximately 10,000x fewer experiences in three of the four Atari games. Although DQN and Blob-PROST outperformed OLRL in Pong, they require over 100x more experiences to acheive the performance of OLRL. Furthermore, SimPLE and SPR, two of the current best agents for Atari in sample-limited regimes, achieved a mean of 54\% and 37\%, respectively, of the human-normalized score across Pong, Asterix, and Breakout after 100,000 steps; our agent achieved a mean of 176\% human normalized score with just 10,000 steps.

\begin{figure}
\centering
\subfloat{
\includegraphics[width=0.25\textwidth]{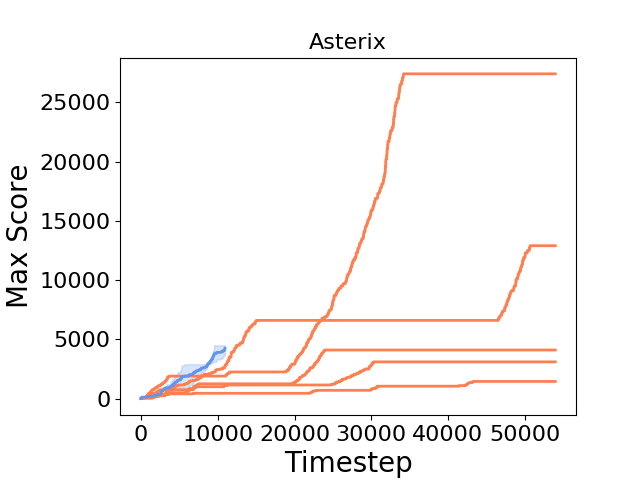} 
}
\subfloat{
\includegraphics[width=0.25\textwidth]{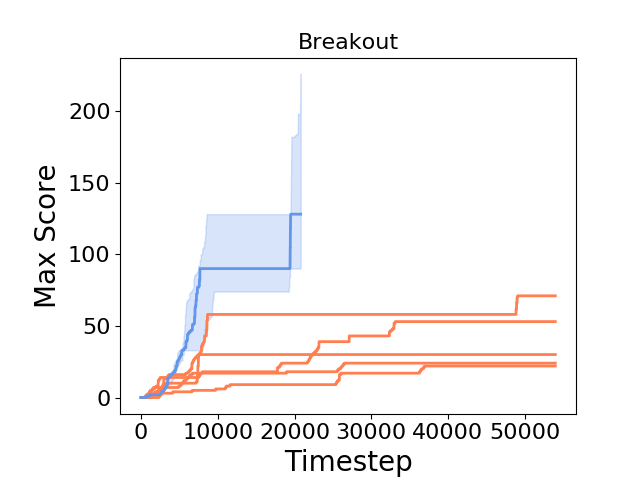} 
}
\\
\subfloat{
\includegraphics[width=0.25\textwidth]{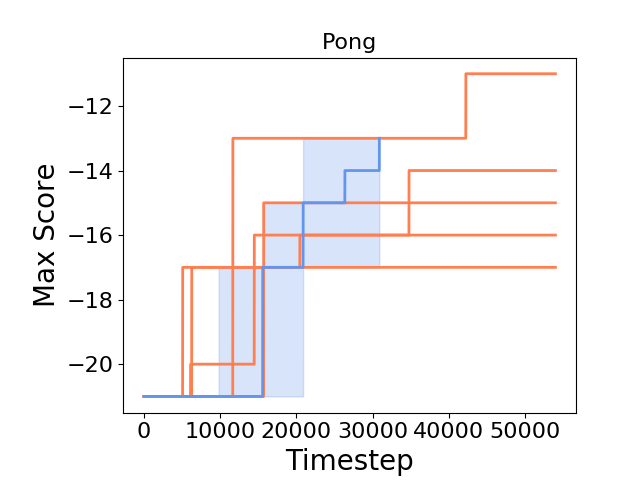}
}
\subfloat{
\includegraphics[width=0.25\textwidth]{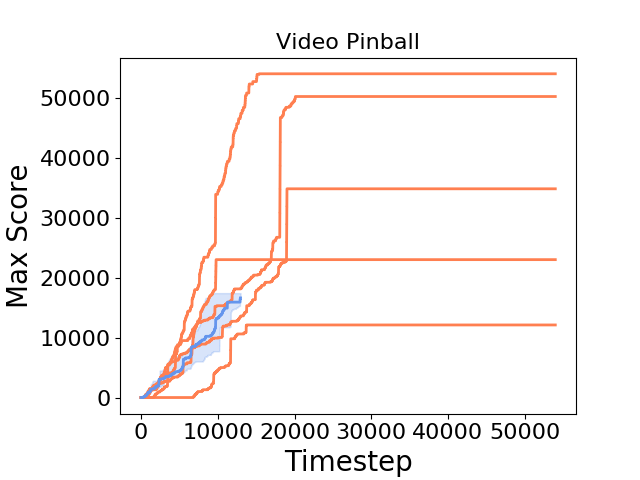}
}
\captionsetup{width=0.5\textwidth}
\captionof{figure}{Comparison of OLRL and human learning curves on Atari games. Each human learning curve is shown in red. Median OLRL score and error bars are shown in blue.}
\label{fig:HumanComp}
\end{figure}

\textbf{Comparison to Humans}: We compared the learning efficiency of OLRL to humans by inviting people to play Atari games and recording the top score they achieved by each timestep. Figure~\ref{fig:HumanComp} compares the maximum scores achieved by each time step for human players and OLRL. For all games tested, this value plateaued for humans, indicating they did not increase their highest score and likely saturated learning. OLRL learns faster than a human if it reaches a human's maximum score in fewer time steps than that human. Despite humans having extensive prior knowledge, OLRL exceeded a human's maximum score in fewer experiences than the human 55\% of the time, demonstrating the power of the object inference and reinforcement learning algorithm we proposed. Details about human play data collection are presented in the appendices.

\section{Conclusion}
Motivated by developmental psychology, we argue in this paper that objects should be defined primarily by their behavioral, rather than just visual, consistency across time. We formalize this notion into an object objective function and optimization algorithm that quickly infers minimal and task-relevant sets of objects with consistent dynamics, value, and reward. We further propose a flexible reward and value estimation architecture that encodes the causality of object contacts. We show that our object-level RL agent learns significantly faster than deep RL algorithms in 2D and 3D environments that feature a variety of realistic object inference challenges. Finally, our OLRL approach is highly modular and adaptable to different segmentation, tracking, reinforcement learning, and sequence modelling algorithms. This paper offers several future avenues of research. At the core of our OLRL agent is an object objective function based on model errors, which could be improved by integrating a more rigorous measurement of model error and uncertainty. In addition, optimizing the object objective function requires expensive training of new dynamics, reward, and value models for each candidate representation and a large memory footprint to store all past experiences. We look forward to our OLRL framework and grounding of objectness in behavior serving as a foundation for further advancements in this area.

\bibliographystyle{named}
\bibliography{references}

\begin{appendices}

\section{Implementation Details}

\subsection{Creating Tracklets}
OORL uses a tracking and segmentation algorithm $\mathbb{S}$ to process observations into tracklets, or tracked image segmentations. We only require that $\mathbb{S}$ oversegment and undertrack, that is, never map two objects onto the same tracklet. OORL then learns a mapping from tracklets to tracks to form a compact and task specific object representation. The oversegmention and undertracking requirement is generally easily met by tuning confidence thresholds; OORL is compatible with a wide variety of segmentation and tracking algorithms. We implement $\mathbb{S}$ by using one of two segmentation algorithms to first oversegment each observation, and then a feature-based tracking algorithm to greedily match successive segmentations into tracklets.

\textit{Segmentation}: For 3D environments, we use UOIS~\cite{xie2019best}, a state of the art deep neural network for unseen object segmentation. For 2D environments, we use Felzenszwalb segmentation \cite{felzenszwalb2004efficient}, a classical segmentation algorithm. Successfully using both a classical segmentation algorithm and a state of the art segmentation DNN demonstrates the flexibility of our framework. For UOIS, we set the \texttt{skip\_pixels} to encourage oversegmentation. For Felzenszwalb segmentation, we use a scale of 1000 and sigma of 0 to encourage oversegmentation.

\textit{Tracking}:
We form tracklets by matching segmentations in successive observations to existing tracklets. Let $L_{-1} $ be the set of existing tracklets in their most recently observed position at time $t$, so each tracklet $l_{-1} \in L_{-1}$ is a $\left\{0,1\right\}^{\times H \times W}$ segmentation. Let $S_{t+1}$ be the segmentations for observations at time $t+1$, where each segmentation $s \in \left\{0,1\right\}^{V \times H \times W}$. Given a tracklet and segmentation, $l_{-1} \in L_{-1}$ and $s_{t+1} \in S_{t+1}$, we assign a scalar tracking error $T(l_{-1}, s_{t+1})$ to the segmentation pair. We then greedily match segmentations to tracklets with the lowest error below a certain threshold $t_e$. Unmatched segmentations are added as new tracklets.

We use a an ensemble of human-inspired tracking features for $T$, detailed in Equation 1.
\small
\begin{equation}
\begin{split}
T(l_{-1}, s_{t+1})=w_{0}\mathcal{F}_{disp}(l_{-1}, s_{t+1})+w_{1}\mathcal{F}_{shape}(l_{-1}, s_{t+1})\\+w_{2}\mathcal{F}_{disp}(l_{-1}, s_{t+1})\mathcal{F}_{shape}(l_{-1}, s_{t+1})+w_{3}\mathcal{F}_{perm}(l_{-1}, s_{t+1})\\+w_{4}\mathcal{F}_{size}(l_{-1}, s_{t+1})+w_{5}\mathcal{F}_{motion}(l_{-1}, s_{t+1})
\end{split}
\end{equation}
\normalsize

$\mathcal{F}_{disp}(l_{-1}, s_{t+1})$ is the total change in relative distance between $l_{-1}$ and $s_{t+1}$ and all objects contacting $s_{t+1}$, capturing the notion that over small changes in time the positions of objects generally do not change too much.
\[\mathcal{F}_{shape}(l_{-1}, s_{t+1})=\sum_{i=1}^{25}|\log(Z_{i}(l_{-1}))-\log(Z_{i}(s_{t+1}))|\] 
where $Z_{i}(o)=i$th Zernike moment (\cite{khotanzad1990invariant}) of $o$, encoding object shape consistency.

$\mathcal{F}_{perm}(l_{-1}, s_{t+1})$ captures the intuition that objects generally do not appear or disappear and is the number of experiences since $s_{t+1}$ was last seen,
\[\mathcal{F}_{size}(l_{-1}, s_{t+1})=\max(\frac{|l_{-1}|}{|s_{t+1}|}, \frac{|s_{t+1}|}{|l_{-1}|})-1. \]
$\mathcal{F}_{motion}(l_{-1}, s_{t+1})$ encodes that many objects are background objects and are unlikely-move:
\[\mathcal{F}_{motion}(l_{-1}, s_{t+1})=\log(\max(P_{motion}(l_{-1}, s_{t+1}), \epsilon)),\]
where $\epsilon$ is some small positive constant.

\tiny
\[P_{motion}(l_{-1}, s_{t+1})=\begin{cases} 
    P_{dm}(\textrm{Moves}(s_{t+1}) | e_{1}, \dots, e_{i}),&  med(l_{-1}) \neq med(s_{t+1})\\
    P_{dm}(\textrm{NotMoves}(s_{t+1}) | e_{1}, \dots, e_{i}),& med(l_{-1}) = med(s_{t+1})
\end{cases} \]
\normalsize

where $med(s)$ is the median coordinates of the observed point cloud composing segment $s$. 

When we merge two tracklets $l,l'$during object inference, we discover an instance where the above segmentation algorithm has failed, as it should have tracket $l$ to $l'$. We can use this information to learn a per-object tracking model. We learn tracking models consisting of two LightGBM \cite{ke2017lightgbm} trees per object. We used LightGBM rather than XGBoost trees because of the large amount of tracking data made the faster training time of LightGBM advantageous. For each object, we trained a lightGBM tree on the negative tracking instances associated with that object and the tracking instance which were once negative but later learned to be positive via merging two tracklets into the object track. We used the tracking features described in Appendix A.1 as inputs. The LightGBM tree was trained to output $1-P_t(l_{-1},s_{t+1})$, where $P_t(l_{-1},s_{t+1})$ is the probability $l_{-1},s_{t+1}$ are the same object. We form the learned tracking algorithm $T_{L}$, by taking the maximum of the base tracking algorithm $T$, and the learned LightGBM tracking algorithm.

\subsection{Dynamics, Value, and Reward Models}
\textit{Dynamics Models}--We model the velocity of each object in each dimension with a separate XGBoost \cite{chen2016xgboost} tree. Each tree outputs a probability vector over velocities, discretized in steps of 1 between -30 and 30. Each tree is trained to minimize multiclass logloss, has a maximum depth of $\max (2, \min(6, |D|/50))$, where $D$ is the size of the training data. Experiences are split into a 80:20 train/validation split, which is used for early stopping of training (5 rounds), and for computing the model likelihoods used in the Infer algorithm in the main paper. Where objects appear when they transition from not being present to being present is modelled with an XGBoost tree per object per dimension trained to output a probability vector over positions. Each tree has a maximum depth of 6 and is trained to minimize multiclass logloss, with a similar 80:20 train/validation data split. Object presence is modelled with an XGBoost tree per object, minimizing multiclass logloss on an 80:20 split with a maximum tree depth of 6. For all models, mean prediction was also evaluated, and if mean prediction yields better validation error it was used instead (we did this primarily to prevent objects with very little associated data from causing noisy validation error behavior in the XGBoost trees and interfering with the Infer algorithm).

\begin{figure}
\centering
\subfloat{
\includegraphics[width=0.4\textwidth]{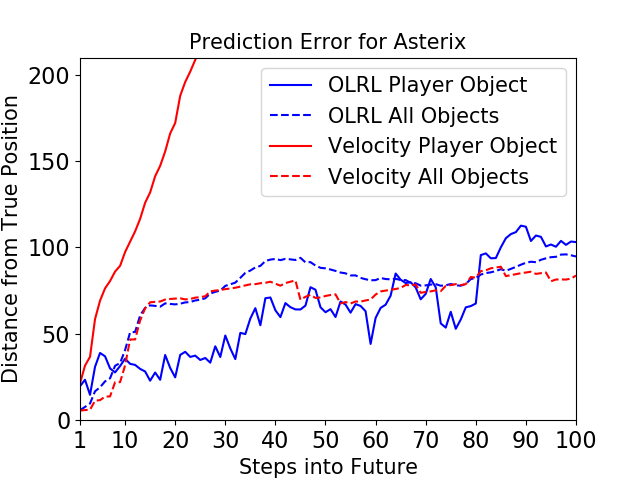}
}
\\
\subfloat{
\includegraphics[width=0.4\textwidth]{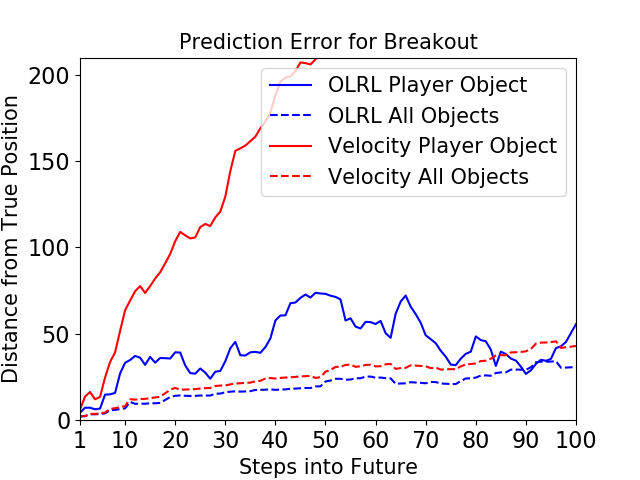}
}
\\
\subfloat{
\includegraphics[width=0.4\textwidth]{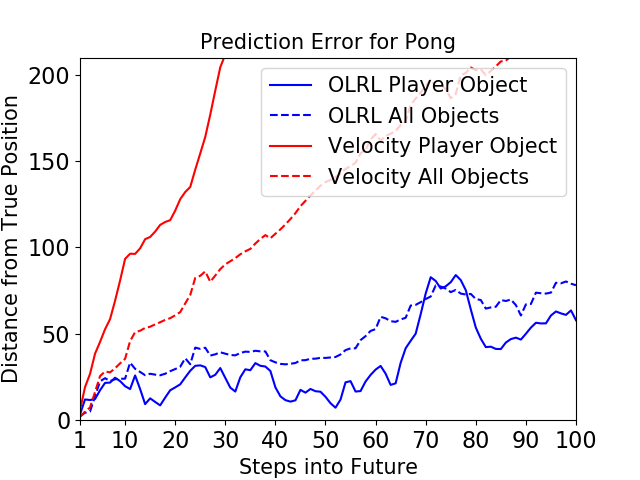}
}
\\
\subfloat{
\includegraphics[width=0.4\textwidth]{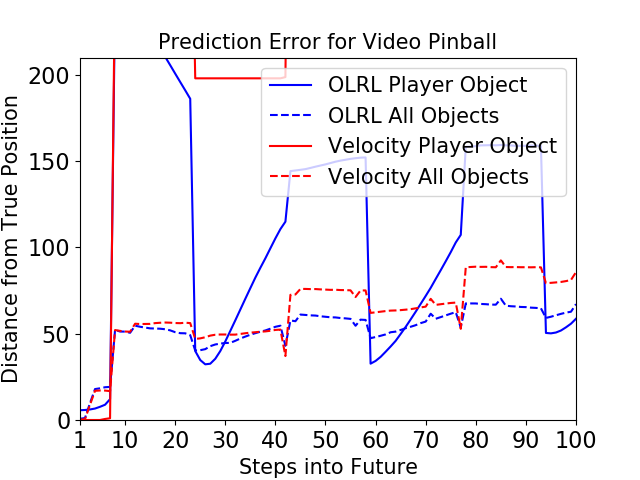}
}
\captionsetup{width=0.4\textwidth}
\captionof{figure}{Model Error on Atari Environments. Distances are in pixels; for comparison, an Atari observation is 210x160 pixels.}
\label{fig:model_error}
\end{figure}

\textbf{Dynamics Model Analysis}: In this section we examine the speed and accuracy of our object-level perception and modeling. For each Atari game, we trained our model for only 2000 observations while playing random actions. We then evaluated our models by first playing $n$ random actions, where $n$ was sampled uniformly from [50, 150], and then predicting the positions of objects for the next 100 steps. Figure \ref{fig:model_error} shows prediction error as the average distance between the predicted and actual object positions. Since many Atari objects do not move, we also show prediction error for the player-controlled object, which has complex dynamics. For the same reason, we use predicting constant object velocity as a natural baseline. Even with very little training data, our models have learned object dynamics with high accuracy tens of frames into the future, effective for short-term planning. In addition, model prediction error generally does not explode for longer prediction horizons, allowing the agent to consider the approximate positions over objects even 100 steps into the future.

\textit{Reward and Value Models}--Our agent uses Monte Carlo Control \cite{sutton1988learning} with a discount factor of 0.95 to estimate state-action values. We then learn to predict state-action reward and value with an XGBoost tree for each object pair. Each XGBoost tree trained to predict deviation form the mean value or reward given the absolute and relative positions, velocities, accelerations, and contact states of the object pair. Each tree was trained to predict the mean of a Gaussian distribution with variance 1 describing the reward distribution for each state-action$\times$object pair and trained to maximize likelihood. Data was split into 3:1 train/validation.

\section{MuJoCo Environments}
In this section we describe the parameters of our 3D MuJoCo environment suite.

\begin{figure}[H]
\begin{center}
\tiny
\begin{tabular}{| c |  c | c |}
\hline
  environment & gather & avoid \\
 \hline
 number target objects & 2 & 2\\ 
 \hline
 object shape & sphere & sphere\\ 
 \hline
 sphere radius & uniform(0.05, 0.1) & uniform(0.05, 0.1)\\
 \hline
 sphere color  & uniform random & uniform random\\
 \hline
 agent speed & 0.1 & 0.1\\
 \hline
 target object speed & 0.2 & 0.05\\
 \hline
 target object contact reward & +1 & -1\\
 \hline
 randomization & none & none\\
 \hline
\end{tabular}
\end{center}
\normalsize
\end{figure}

\begin{figure}[H]
\begin{center}
\tiny
\begin{tabular}{| c |  c | c | c | c |}
\hline
  environment & gather+r & avoid+r \\
 \hline
 number target objects & 2 & 2\\ 
 \hline
 object shape & sphere & sphere\\ 
 \hline
 sphere radius & uniform(0.05, 0.1) & uniform(0.05, 0.1)\\
 \hline
 sphere color &  uniform random & uniform random \\
 \hline
 agent speed & 0.1 & 0.1\\
 \hline
 target object speed & 0.2 & 0.05\\
 \hline
 target object contact reward & +1 & -1\\
 \hline
 randomization & color+size & color+size\\
 \hline
\end{tabular}
\end{center}
\normalsize
\end{figure}

\section{Human Atari Learning Data}
We conducted a study of human Atari game play for four games to obtain learning curves. Five participants were tasked with playing Asterix, Breakout, Pong, and Video Pinball for 15 minutes each. To ensure the environment the humans were tested on was identical to the environment we trained our agents on, we used OpenAI Gym \cite{openai_gym} as the Atari emulator. We found that there is very high variance in score between participants, even on Asterix, which participants were unlikely to have played before. In every game at least one participant was able to beat the expert score in 15 minutes of play.

\section{Computing Infrastructure and Reproducability}
The experiments in this paper were run on one GPU, 8 cpu cores, and 128GB of RAM. Each experiment took approximately one hour to complete, which we note is very reasonable for deep reinforcement learning.

\section{Ethics Statement}
While this work is not immediately applicable to real-world or societal problems, it advances a reinforcement learning paradigm that is much more efficient than existing architectures, lowering compuational/resource barriers to research, and is very intuitive to humans, opening the way for research into interpretable and explainable learners and planners.

\end{appendices}

\end{document}